# Neural Network-Based Modeling of Hysteretic Systems: From Prandtl and Preisach Neural Networks to Neuro-Madelung and Neuroplastic Continuous Models

Mehrdad Shafiei Dizaji

Postdoctoral Research Associate,

Civil Engineering, University of Virginia, U.S.

Farzad Shafiei Dizaji

Researcher

Civil Engineering, University of Virginia, U.S.

## Abstract

Hysteresis is a complex nonlinear phenomenon characterized by memory dependence, path dependence, and, in many physical systems, degradation, non-congruency, asymmetry, and rate dependence. These characteristics pose significant challenges for conventional artificial neural networks because standard feedforward architectures are inherently memoryless, while recurrent and autoregressive approaches generally represent memory through previous input-output states rather than through mechanisms specifically associated with hysteresis. Over the past two decades, several specialized neural-network architectures have therefore been developed by incorporating concepts from classical hysteresis theory directly into neural computation. This review examines the development and capabilities of Prandtl neural networks (PNNs), Preisach neural networks (Preisach-NNs), generalized Prandtl neural networks (GPNNs), the neuro-Madelung model (NMM), extended Preisach neural networks (EPNNs), and dynamic plastic continuous neural networks (DPCNNs). Particular attention is given to their mechanisms for representing memory, network architectures, activation functions, training algorithms, treatment of deterioration and rate dependence, and applications to nonlinear structural and material systems. The PNN introduced hysteretic memory directly into neural activation through stop operators and demonstrated accurate nonlinear dynamic analysis of structures. Preisach-NNs subsequently extended this concept through an additional nonlinear hidden layer to represent non-Masing hysteresis. GPNNs introduced deteriorating-stop neurons to capture stiffness and strength degradation and non-congruent hysteresis. A different generalization was provided by the NMM, which combined Madelung's hysteresis rules with multilayer feedforward neural networks to represent diverse rate-independent and rate-dependent behaviors. These developments culminated in the EPNN, which combines deteriorating hysteretic operators, nonlinear neural mappings, and input-rate information to accommodate Masing and non-Masing, symmetric and asymmetric, congruent and non-congruent, and rate-independent and rate-dependent hysteresis. In parallel, DPCNNs introduced a fundamentally different paradigm in which neurons interact through an adaptive nonlinear plastic continuum, allowing wave propagation and material plasticity to contribute to information processing and memory. Collectively, these studies demonstrate a transition from neural networks that merely approximate hysteretic input-output relations toward architectures in which physical mechanisms of memory, plasticity, deterioration, and rate dependence are embedded directly into the computational model. The review identifies the strengths and limitations of these approaches and discusses opportunities for developing physics-informed, interpretable, and computationally efficient neural models for complex hysteretic systems.

## 1. Introduction

Hysteresis appears in a broad range of engineering and physical systems, including structural materials, ferromagnetic and ferroelectric materials, piezoelectric and magnetostrictive actuators, shape-memory alloys, and nonlinear mechanical systems. Unlike a single-valued nonlinear relationship, the output of a hysteretic system cannot generally be determined from the instantaneous input alone. Instead, its response depends on the loading history and the internal state accumulated during previous loading cycles. Classical models such as the Masing, Prandtl-Ishlinskii, Preisach, and Bouc-Wen formulations have consequently played important roles in describing hysteretic phenomena.

The principal difficulty with classical mathematical models is that no single prescribed mathematical form is universally suitable for materials and systems exhibiting substantially different hysteretic characteristics. This limitation motivated the application of artificial neural networks as model-free nonlinear approximators. However, conventional multilayer feedforward neural networks are static and possess no intrinsic memory. Approaches such as NARX, recurrent, and time-delay neural networks introduce historical information, but their representation of hysteretic memory may require previous responses or internal variables and can consequently encounter error accumulation during recursive prediction (Farrokh et al., 2022).

A different research direction emerged by incorporating the physics and mathematical operators of hysteresis directly into neural-network architectures. The Prandtl neural network (PNN), for example, incorporates stop operators into its hidden neurons. Because these operators themselves possess hysteretic memory, the network does not require the complete past response to be explicitly supplied to its input layer (Joghataie and Farrokh, 2008).

This idea subsequently developed in several directions. Preisach-NNs introduced a second hidden layer of sigmoidal neurons, extending the capability of the PNN from Masing to more complicated non-Masing behavior. GPNNs replaced conventional stop neurons with deteriorating-stop neurons to represent stiffness and strength degradation. The NMM approached generalization differently, using Madelung's rules to organize hysteretic memory while assigning the remaining nonlinear mapping to a multilayer neural network. More recently, the EPNN combined several of these concepts and incorporated input-rate information to address rate-dependent hysteresis (Farrokh and Joghataie, 2014).

In parallel with these operator-based architectures, DPCNNs proposed a radically different concept: neurons communicate through a continuous nonlinear plastic medium rather than solely through conventional wire-like connections. Information propagates through this medium as waves, and its plastic properties provide an additional mechanism for learning and memory (Joghataie and Oliyan

Torghabehi, 2014; Joghataie and Shafiei Dizaji, 2016). This review synthesizes these developments and evaluates them from four perspectives: (1) representation of hysteretic memory, (2) network architecture, (3) training and adaptation, and (4) ability to reproduce increasingly complicated hysteretic characteristics.

## 2. Why Conventional Neural Networks Struggle with Hysteresis

The central computational difficulty associated with hysteresis is its non-unique input-output relationship. The same instantaneous input may correspond to different outputs depending on the previous loading trajectory. Consequently, a conventional feedforward mapping is generally insufficient.

$$y(t) = F[x(t)]$$

A network must instead contain or receive information describing the system history. Conventional approaches accomplish this by augmenting the input vector with previous states.

$$y(t) = F[x(t), x(t-1), y(t-1), ...]$$

Operator-based neural networks offer a conceptually different solution: put the memory inside the neuron itself. In a PNN, the hidden-layer activation is a stop operator rather than a conventional sigmoid or rectified activation function. The stop operator has its own state and therefore remembers the loading path. A Prandtl–Ishlinskii stop operator is schematically shown in Figure 1.

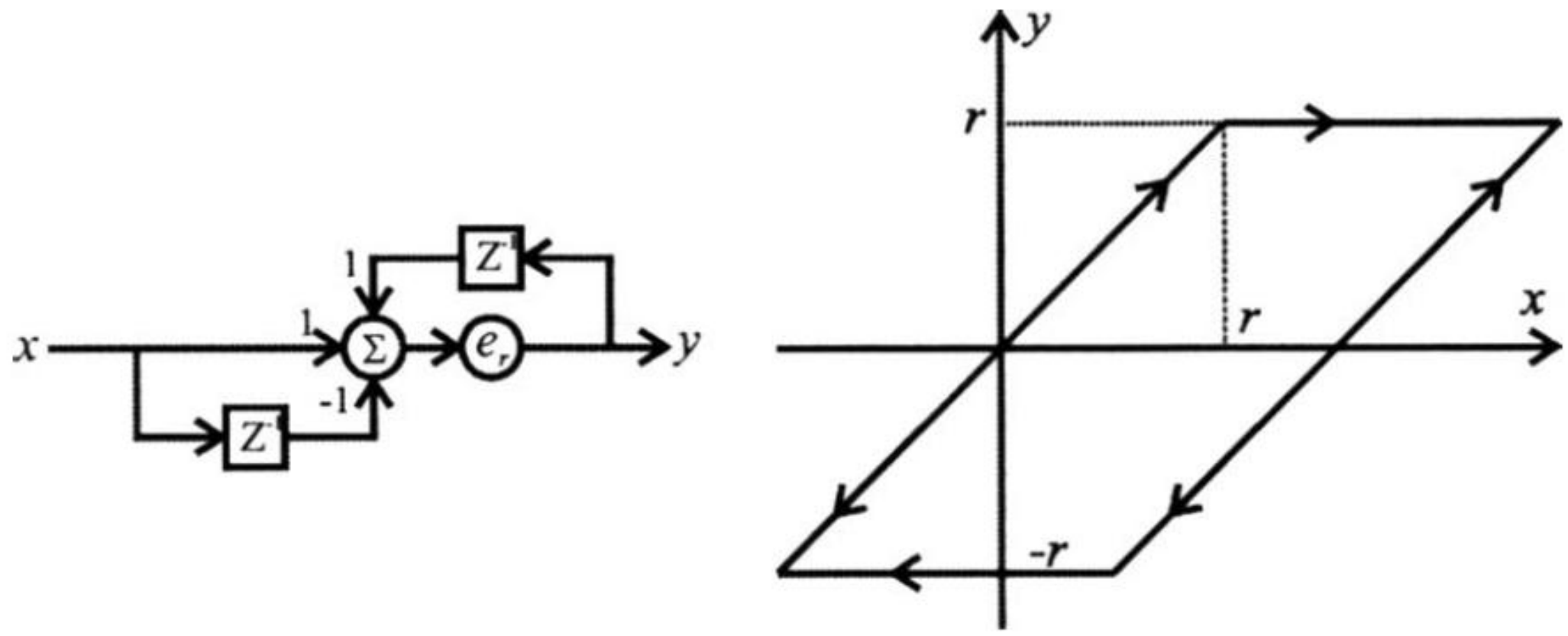

*Figure 1. (a) Stop operator, (b) Hysteresis loop of stop operator*

## 3. Prandtl Neural Network: Embedding Hysteresis into Neurons

PNN constitutes the foundation of several subsequent developments. It was inspired by the Prandtl-Ishlinskii representation, in which a hysteretic response is constructed as a weighted superposition of stop operators. Accordingly, a PNN contains linear input and output neurons and a hidden layer composed of stop neurons (Joghataie and Farrokh, 2008). The important distinction between a stop neuron and a conventional neuron is that its activation is history dependent. Consequently, hysteretic memory is embedded locally within the network.

Training of the original PNN was formulated as an optimization problem involving connection weights and stop-operator yield points. Because the latter create a nonsmooth optimization problem, a continuous genetic algorithm was employed. The genetic algorithm operates directly on the objective function without requiring its derivatives (Joghataie and Farrokh, 2008).

The model was demonstrated using nonlinear one- and three-story shear frames subjected to earthquake excitation. Training was performed using white-noise excitation, whereas earthquake records not included in training, including the 200% El Centro and Gilroy records, were employed for testing. The PNN responses closely reproduced numerical-integration solutions even for strongly nonlinear hysteretic behavior (Joghataie and Farrokh, 2008).

The PNN therefore established an important principle: hysteresis-specific activation functions can provide neural networks with intrinsic memory. Its principal limitation, however, follows directly from the conventional stop operator. Stop neurons generate elastic-perfectly plastic, congruent behavior and do not inherently represent deterioration (Joghataie and Farrokh, 2008; Farrokh et al., 2015). Figure 2 shows the architecture of a Prandtl neural network. It is a FFNN with three layers: an input and an output layer, each with one linear unit representing x(t) and y(t), respectively, and only one hidden layer which is comprised of *m* stop operators as the hidden neurons

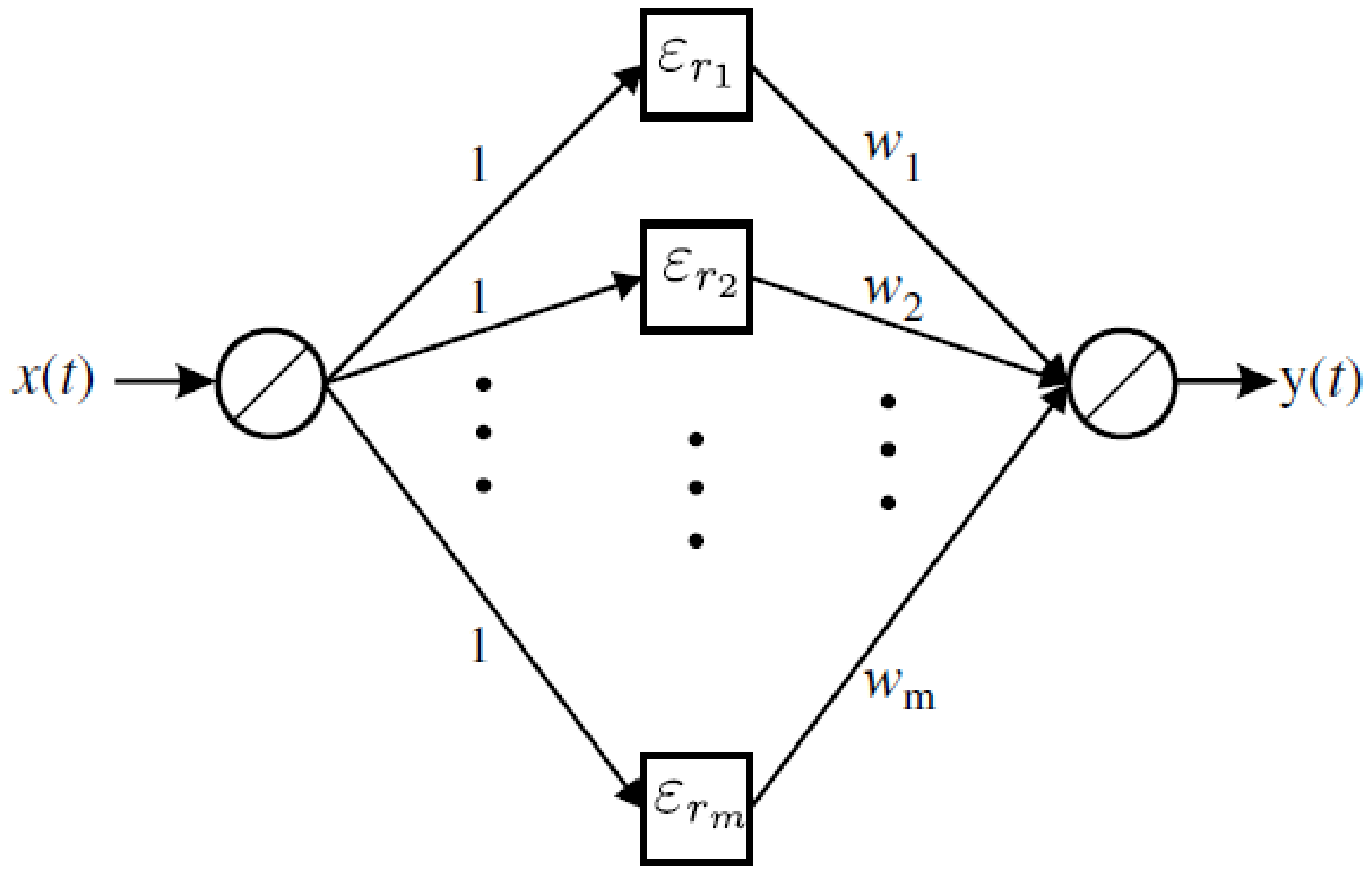

*Figure 2. Architecture of Prandtl neural network*

### 4. Preisach Neural Network: Extending PNNs to Non-Masing Hysteresis

The Preisach-NN was developed to increase the flexibility of the PNN while retaining operator-based memory. It contains two hidden layers. The first contains stop neurons, while the second contains conventional sigmoidal neurons (Farrokh and Joghataie, 2014). This additional nonlinear mapping is crucial. Whereas PNN is essentially a linear combination of stop-operator outputs, the second hidden layer allows nonlinear combinations of these memory states.

The studies considered both Masing and non-Masing hysteresis. For Masing behavior, the sigmoidal layer was unnecessary, and the architecture effectively reduced to a Prandtl-type network. For non-Masing hysteresis, however, the sigmoidal neurons were necessary. The reported mean square error for one non-Masing case was approximately $2.41 \times 10^{-5}$, indicating highly accurate identification (Farrokh and Joghataie, 2014). Training also changed significantly. Instead of relying exclusively on a genetic algorithm, the Preisach-NN employed a subgradient optimization method with space dilatation, appropriate for its nonsmooth optimization problem (Farrokh and Joghataie, 2014). Details of Preisach-NN can be seen in Figure 3. Nevertheless, the model inherited important characteristics of the Preisach framework. In particular, it remained fundamentally suited to congruent and rate-independent hysteresis and therefore could not provide a complete treatment of non-congruency or rate dependence (Farrokh and Joghataie, 2014; Farrokh et al., 2022).

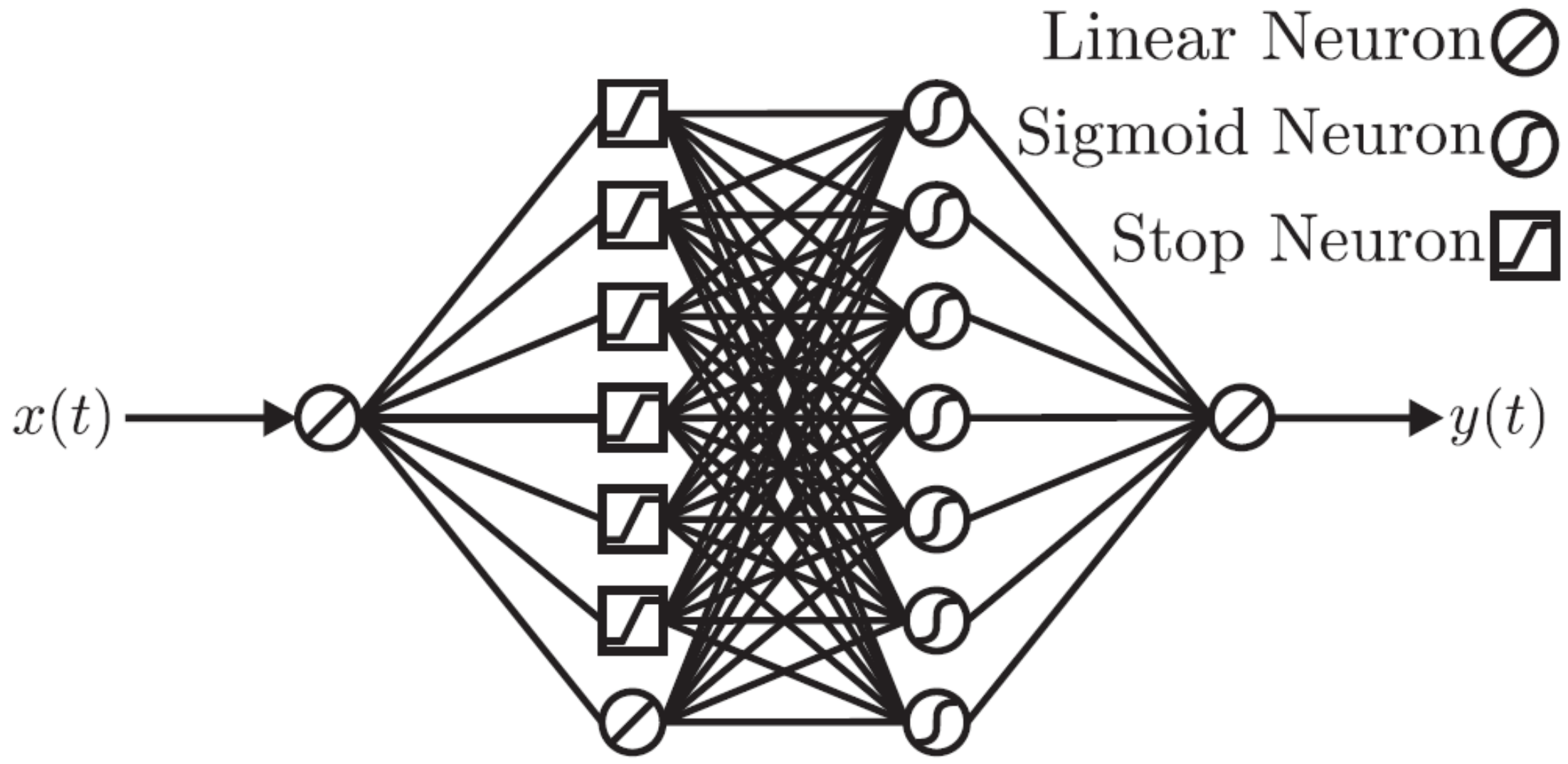

*Figure 3. Preisach-NN architecture*

**5. Generalized Prandtl Neural Network: Introducing Damage and Deterioration**

The GPNN addressed another major deficiency of the original PNN: its inability to represent material deterioration (Farrokh et al., 2015). Damage during cyclic loading may cause both strength degradation and stiffness degradation. Consequently, loops having identical input extrema may no longer have identical shapes. This violates the congruency assumption underlying many classical hysteresis models. To address this issue, a deteriorating stop (DS) operator was developed from a combination of stop and play operators. Cumulative slip is employed as an indicator of accumulated damage, while a deterioration parameter controls the evolution of the operator. Decreasing this parameter increases both stiffness and strength degradation; for sufficiently large values, the DS operator approaches the ordinary stop operator (Farrokh et al., 2015). Replacing the PNN's stop neurons with DS neurons produces the GPNN. The overall architecture remains closely related to the PNN, but the memory elements now evolve with accumulated damage (Farrokh et al., 2015).

The GPNN was evaluated using mathematical hysteresis models with and without deterioration, experimental data exhibiting severe nonlinear damage, and dynamic analysis of a deteriorating one-story shear frame. The model successfully reproduced experimental hysteresis even when testing involved an input range exceeding that used during training (Farrokh et al., 2015). In the structural-dynamics example, the GPNN neuro-modeler also outperformed the earlier PNN when severe damage was introduced (Farrokh et al., 2015).

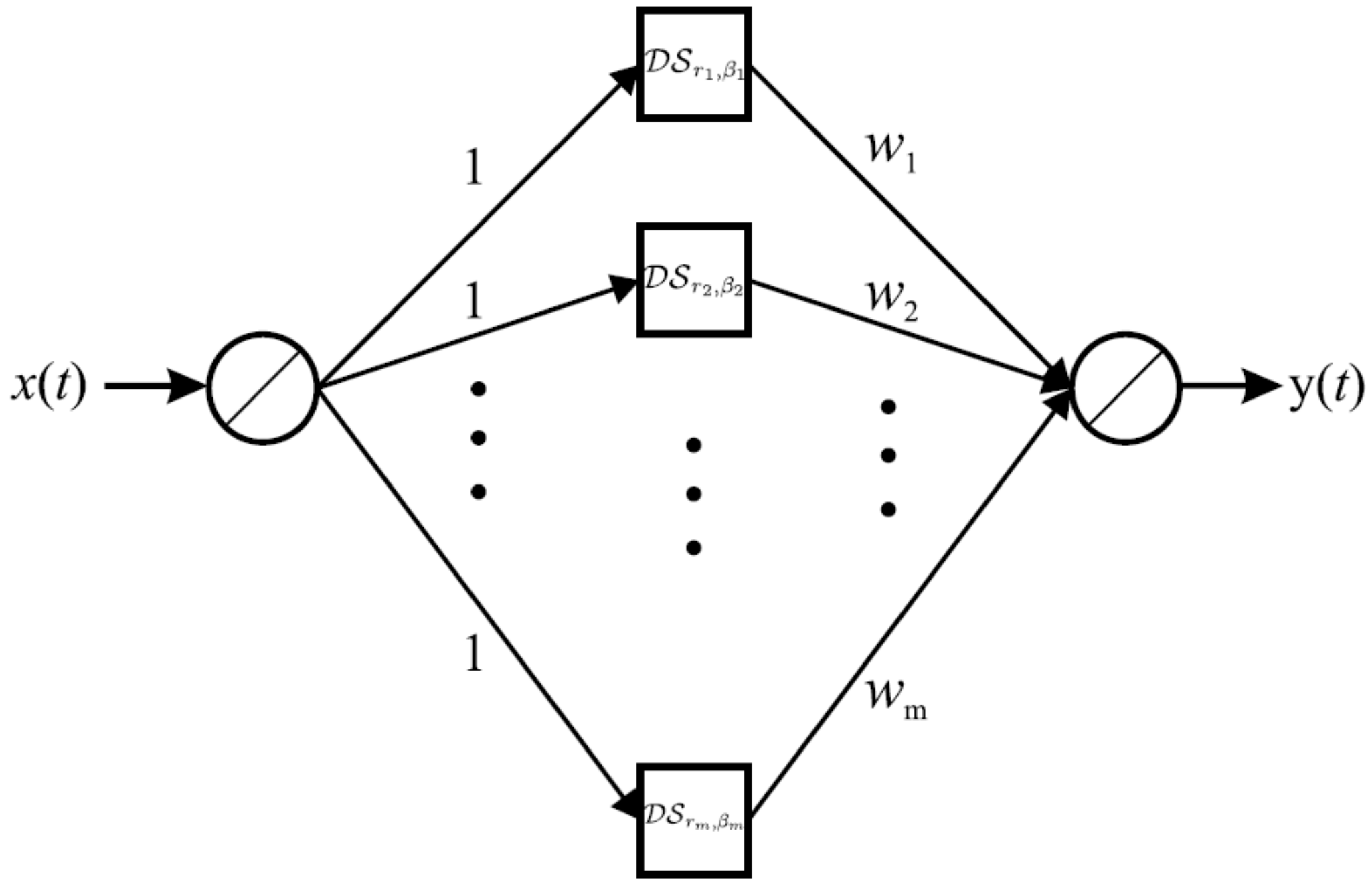

*Figure 4. Generalized Prandtl neural network architecture*

**6. Neuro-Madelung Model: A Rule-Based Generalization**

The NMM represents a substantially different approach (Farrokh and Dizaji, 2016). Rather than deriving the network directly from a specific superposition of Prandtl or Preisach operators, the model begins with Madelung's rules, particularly their representation of turning points and the wiping-out property of hysteretic memory. These rules describe how nested loops are created and erased during changes in loading direction (Farrokh and Dizaji, 2016).

There are three rules: a) any curve G1 emanating from a turning point A in Figure 5 is uniquely determined by the coordinates of point A. b) If any point B on the curve G1 becomes a new turning point, then the curve G2 originating as B leads back to point A. c) If the curve G2 is continued beyond the point A, then it coincides with the continuation of the curve G which led to the point A before the G1–G2-cycle was traversed.

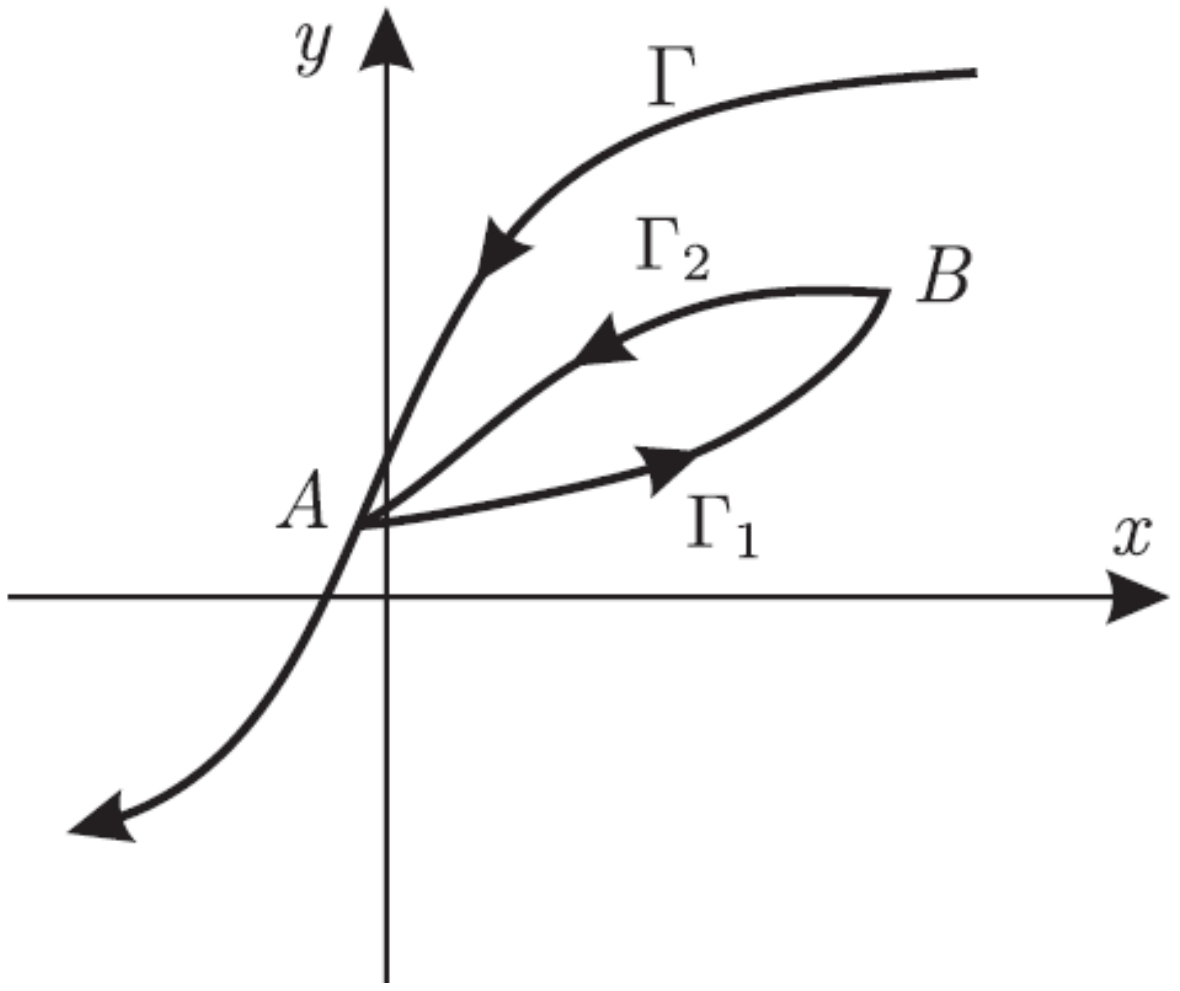

*Figure 5. Deletion property according to Madelung's rules*

The nonlinear static relationship that remains after accounting for this memory structure is then represented using a multilayer feedforward neural network. The input can include the current input, the relevant previous turning-point input-output pair, and the input rate for rate-dependent behavior.

A fixed architecture with two hidden layers containing 15 sigmoid neurons each was used across the investigated examples, and the connection weights were trained through backpropagation using the Levenberg-Marquardt method (Farrokh and Dizaji, 2016). This construction allowed the same basic architecture to address rate-independent non-congruent, asymmetric, and rate-dependent hysteresis. The reported simulation errors were low, and comparisons with alternative models generally favored the NMM.

Another noteworthy contribution was formulation of an implicit inverse NMM. The inverse was investigated for open-loop compensation of rate-dependent hysteresis, with satisfactory preliminary results (Farrokh and Dizaji, 2016). The NMM therefore demonstrates an alternative route toward generality: explicit hysteresis rules for memory combined with a neural network for the unknown nonlinear mapping. The schematic architecture of the considered MLFFNN in this article is shown in Figure 6.

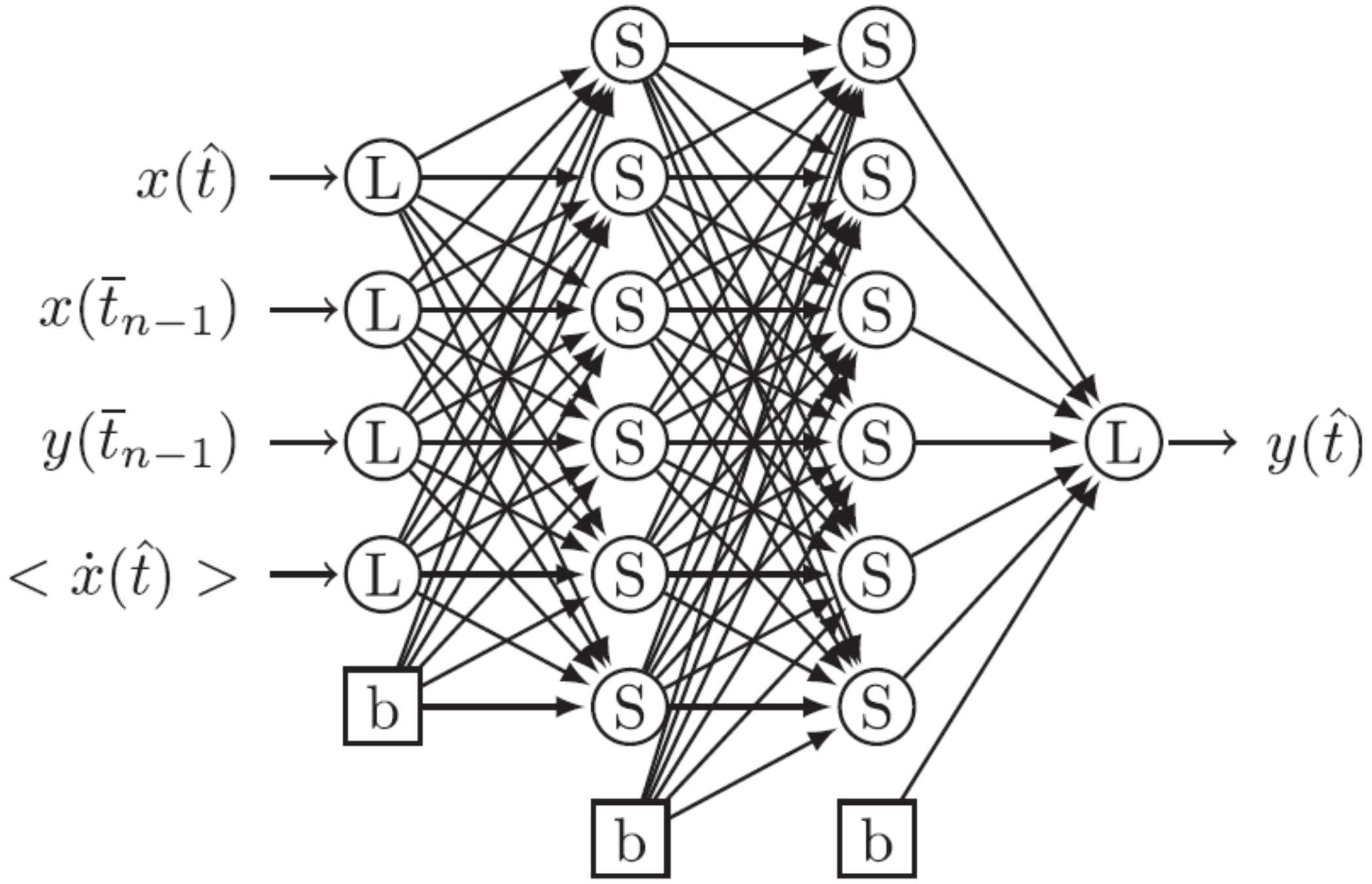

*Figure 6. Schematic of the architecture of the MLFFNN*

**7. Extended Preisach Neural Network: Toward a Unified Hysteresis Model**

The EPNN can be interpreted as an effort to consolidate several developments of the PNN family (Farrokh et al., 2022). The limitations preceding it can be summarized as follows: the PNN is mainly suited to Masing, symmetric, congruent, rate-independent behavior; the GPNN adds non-congruency and deterioration but remains limited regarding rate dependence and general asymmetric non-Masing behavior; and the Preisach-NN addresses non-Masing behavior but retains restrictions associated with congruency and rate independence. The EPNN combines these extensions in a single architecture. It uses two hidden layers, with specialized deteriorating or decaying stop-type neurons in the first and sigmoidal neurons in the second. Crucially, the input vector includes both the input x(t) and its rate x_dot(t), allowing input rate to influence the hysteretic state (Farrokh et al., 2022).

Details of EPNN architecture have been shown in Figure 7. Obviously, the architecture of the EPNN has been considered. It encompasses all specifications of the previously proposed PNN, GPNN, and Preisach-NN in a unique structure. Furthermore, it is a rate-dependent neural network, while neither of the aforementioned neural networks could be utilized for the rate-dependent cases.

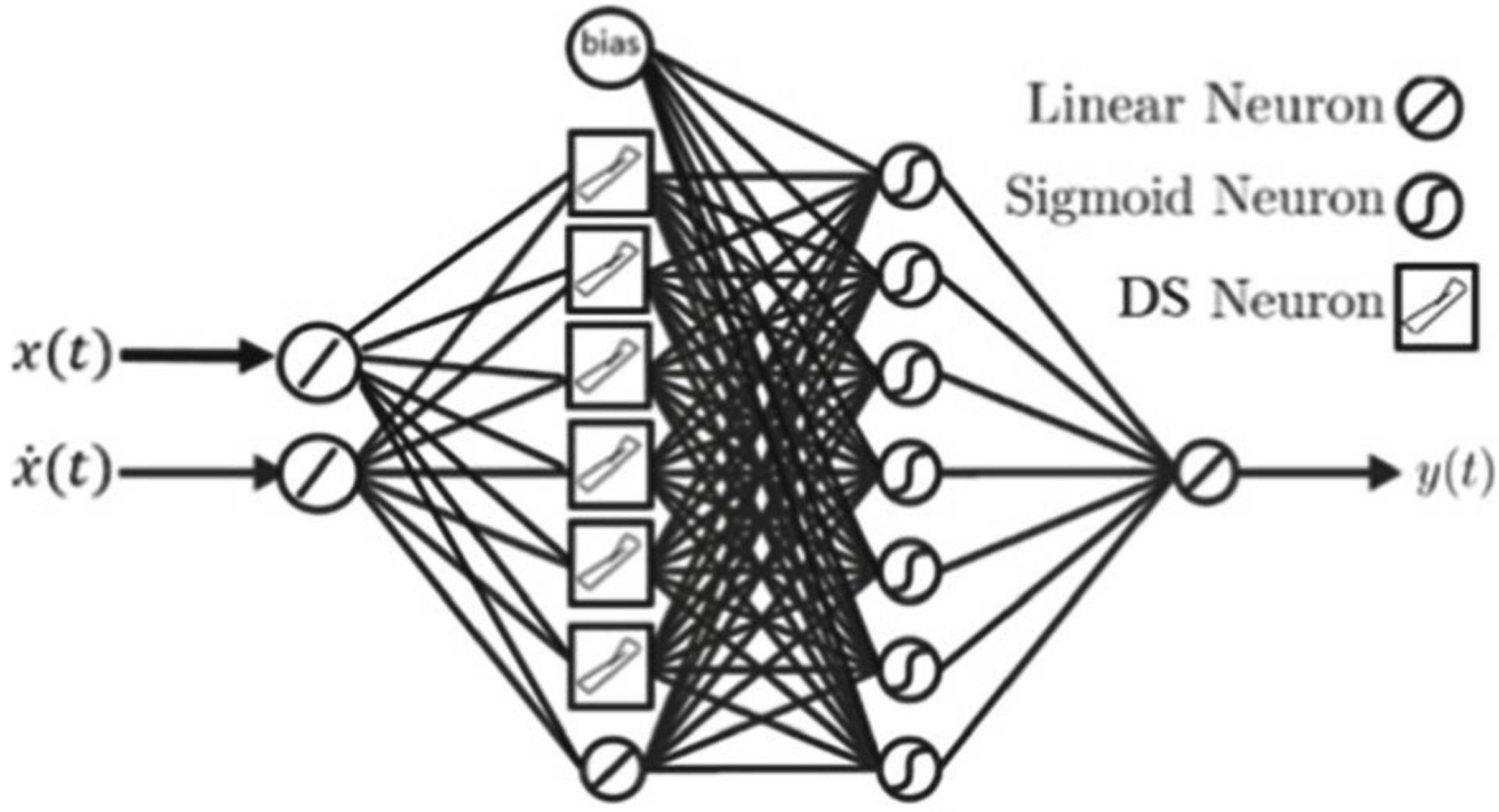

Figure 7. Extended preisach neural network architecture

As a result, the EPNN was designed to accommodate Masing and non-Masing behavior; symmetric and asymmetric loops; congruent and non-congruent hysteresis; and rate-independent and rate-dependent response. Training was performed using a hybrid algorithm combining the advantages of the genetic algorithm and subgradient optimization with space dilatation. The genetic algorithm provides a mechanism for locating a favorable region of the parameter space, while subgradient optimization accelerates subsequent convergence (Farrokh et al., 2022).

The importance of the rate input was demonstrated particularly clearly in a rate-dependent example. An EPNN excluding the input-rate term produced a very large error, whereas incorporating the rate term reduced the reported error to approximately 7 × 10^-5. Testing on a different frequency distribution resulted in an MSE of approximately 0.002 (Farrokh et al., 2022). Experimental examples also included concrete and smart-material behavior. For cyclic concrete, the EPNN was able to represent asymmetric, non-Masing, non-congruent hysteresis that the preceding PNN-family models could not simultaneously accommodate (Farrokh et al., 2022).

## 8. Dynamic Plastic Continuous Neural Networks: A Different Concept of Memory

The two DPCNN papers constitute a related but conceptually distinct research branch.

Rather than representing neural computation exclusively through discrete nodes connected by scalar weights, the DPCNN introduces a continuous nonlinear plastic medium through which information propagates (Joghataie and Oliyan Torghabehi, 2014). In the original formulation, neurons are distributed in or connected to a nonlinear plastic medium, and the finite element method is used to simulate information propagation dynamically. The properties of the medium become trainable variables. This produces an unusual interpretation of neural memory. Memory is not restricted to the internal state of an artificial neuron; it can be stored physically or mathematically in the plastic state of the connecting medium. The original numerical DPCNN was trained by optimizing mechanical properties of the medium and neural capacities using sensitivity analysis and steepest descent (Joghataie and Oliyan Torghabehi, 2014).

The subsequent neuroplasticity study expanded this idea by making both synaptic connection weights and material properties adaptive. The nonlinear plastic base medium contributes to remembering the history of the input-response relationship, motivating the analogy with neuroplasticity (Joghataie and Shafiei Dizaji, 2016). Training was divided into optimization of neural connection weights followed by optimization of selected properties of the base material (Joghataie and Shafiei Dizaji, 2016).

DPCNN general characteristics and architecture as discussed in Joghataie and Olian Torghabehi (2014): As represented in Figure 8(a), a continuous neural network in this paper is defined as comprised of a number of neurons and a plate that forms the base which supports the neurons. Each neuron is connected to the base by wire-like connections. Similar to perceptrons, each neuron has several input connections which take the information to the neuron, a body which is in charge of processing the information to issue an output signal, and a number of wires which transfer and feedback the output signal to the plate. The plate is the medium through which the information is propagated and shared between the neurons. In addition, since the medium is plastic, it contributes to the information retention in the neural networks too and hence is considered as a part of the memory. In Joghataie and Olian Torghabehi (2014) the continuous neural network was called Dynamic Plastic Continuous Neural Network (DPCNN). In general, the neural network can be 3-D where the neurons are either submerged in the medium and/or installed on the surrounding surface of the medium as shown in Figure 8(b).

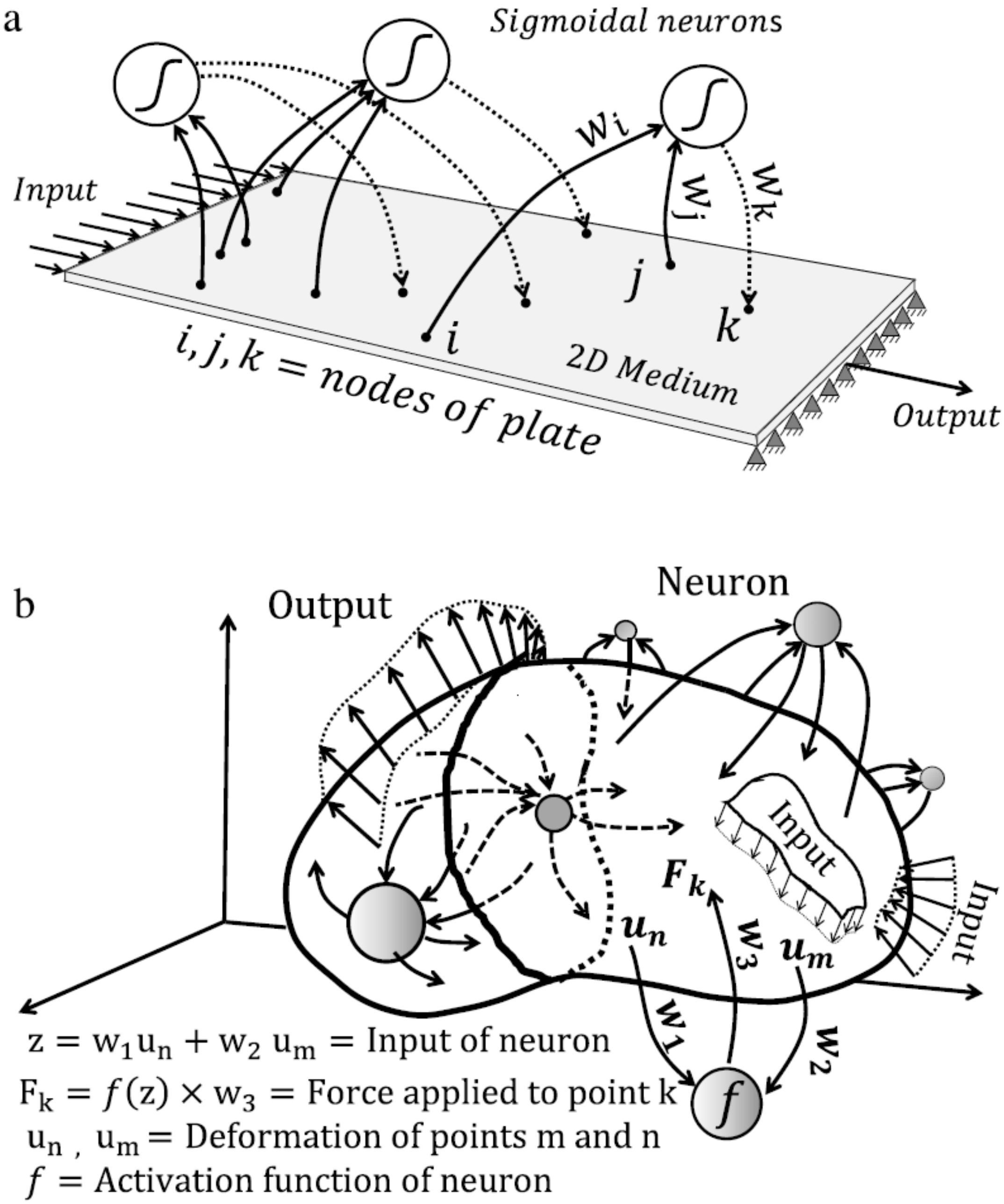

*Figure 8. DPCNNs. (a) 2-D Dynamic Plastic Continuous Neural Network (DPCNN) with neurons on plate medium by input and output connections., (b) General 3-D DPCNN.*

The resulting architecture therefore provides two interacting learning mechanisms: synaptic adaptation and material/medium adaptation. This is fundamentally different from PNN-type networks because the memory mechanism emerges partly from plastic deformation and wave propagation in a spatially continuous computational medium.

## 9. Comparative Evolution of the Models

The seven papers collectively reveal two related trajectories. The operator-based family progressively expands hysteretic capability from intrinsic memory toward deterioration, non-Masing behavior, non-congruency, asymmetry, and rate dependence, while the DPCNN family explores a spatially continuous and material-based form of learning and memory.

Table 1. Comparison of NN models for hysteresis behavior simulation

| **Model** | **Main source of memory** | **Major advance** | **Main limitation emphasized by subsequent work** |
|---|---|---|---|
| PNN (Joghataie and Farrokh, 2008) | Stop operators | Intrinsic hysteretic neural memory | Limited for deterioration, non-congruency, and rate dependence |
| Preisach-NN (Farrokh and Joghataie, 2014) | Stop neurons + nonlinear sigmoid mapping | Non-Masing capability | Congruency and rate-independence restrictions |
| GPNN (Farrokh et al., 2015) | Deteriorating-stop operators | Strength/stiffness degradation and non-congruency | Limited general non-Masing/asymmetric and rate-dependent capability |
| NMM (Farrokh and Dizaji, 2016) | Madelung turning-point memory + MLFFNN | Broad, rule-based representation including rate dependence | Requires explicit turning-point/history management |
| EPNN (Farrokh et al., 2022) | Deteriorating/decaying stop neurons + nonlinear mapping + input rate | Unified treatment of several major hysteresis properties | Increased architecture/training complexity |
| DPCNN (Joghataie and Oliyan Torghabehi, 2014; Joghataie and Shafiei Dizaji, 2016) | Plastic continuous medium + dynamic propagation | Spatial, material-based learning and memory | Computationally intensive finite-element-based framework and early-stage theoretical implementation |

## 10. Evolution of Training Strategies

Another interesting pattern across these studies is the evolution of optimization methods.

The PNN and GPNN rely heavily on genetic algorithms, which are suitable for the nonsmooth optimization created by hysteretic activation functions. The Preisach-NN instead employs subgradient optimization with space dilatation because its larger number of connection weights makes exclusive genetic-algorithm optimization increasingly expensive. The EPNN subsequently combines these ideas into a hybrid algorithm. Genetic optimization addresses the difficulty of choosing a suitable starting point, whereas the subgradient method provides faster local convergence. The NMM avoids this difficulty because its nonlinear approximation is performed using conventional differentiable sigmoid neurons, permitting Levenberg-Marquardt backpropagation. DPCNN training is different again: gradients are obtained through sensitivity analysis of the finite-element representation, and both neural and material variables can be optimized. Thus, the selected training algorithm is closely connected to the physical and mathematical structure used to represent memory.

## 11. Critical Discussion and Research Gaps

A major strength of these studies is their movement away from treating neural networks as generic black-box approximators. Instead, prior knowledge about hysteresis is embedded into the network through stop operators, deterioration laws, Madelung rules, plasticity, or continuous media. This produces what can be regarded as an early form of physics-guided neural modeling. The physical knowledge does not appear merely as a penalty in the loss function; it influences the architecture itself. Nevertheless, several research opportunities remain. First, increased hysteretic generality generally increases computational complexity. The progression from PNN to EPNN requires additional hidden layers, specialized parameters, rate information, and more sophisticated optimization. Efficient differentiable formulations of hysteretic operators could therefore be valuable.

Second, most of these approaches were developed before the recent expansion of modern deep-learning and physics-informed learning techniques. Integrating their physically meaningful memory mechanisms with automatic differentiation, recurrent state-space learning, neural operators, or physics-informed neural networks represents a promising direction.

Third, uncertainty quantification remains largely unexplored in the uploaded studies. Experimental hysteresis may vary because of measurement uncertainty, material variability, loading history, temperature, aging, and damage. Probabilistic extensions could therefore provide confidence intervals rather than deterministic predictions alone. Fourth, transferability across materials deserves further study. The EPNN and NMM demonstrate broad capabilities across different hysteresis classes, but systematic transfer learning between structural materials, smart actuators, magnetic materials, and other systems could substantially reduce retraining requirements.

Finally, the DPCNN concept raises an especially interesting research question: whether plasticity and spatial propagation can be exploited computationally without performing a full nonlinear finite-element simulation during every neural evaluation. Reduced-order models or differentiable continuum representations could make this concept considerably more computationally practical.

## 12. Conclusions

The reviewed papers demonstrate a progressive development of neural networks specifically designed for systems with memory and hysteresis. The PNN established the central concept of embedding hysteretic memory directly into neural activation functions through stop operators. It successfully modeled nonlinear structural dynamics and avoided relying solely on conventional neural-network memory mechanisms. The Preisach-NN introduced a nonlinear second hidden layer, significantly improving the representation of non-Masing hysteresis. The GPNN addressed a different limitation by introducing deteriorating-stop neurons capable of reproducing stiffness and strength degradation and non-congruent behavior. The NMM pursued generality through a different strategy, separating hysteretic memory governed by Madelung's rules from an adaptive neural representation of the remaining nonlinear mapping. This allowed one general framework to represent rate-independent and rate-dependent hysteresis and also supported formulation of an inverse model for control. The EPNN represents the broadest operator-based model among the reviewed studies. By combining deteriorating memory neurons, nonlinear hidden-layer mapping, and input-rate information, it was designed to reproduce Masing/non-Masing, symmetric/asymmetric, congruent/non-congruent, and rate-independent/rate-dependent hysteresis. Its reported agreement with experimental and numerical data demonstrates the value of systematically embedding hysteresis characteristics into neural architecture. Finally, DPCNNs provide an alternative conception of neural computation in which information travels through a nonlinear plastic continuum. Their adaptive base medium allows material properties themselves to participate in learning and memory, providing a connection between artificial neural computation, continuum mechanics, and neuroplasticity.

Taken together, these studies suggest an important broader direction for computational mechanics: rather than asking a generic neural network to discover all aspects of complex nonlinear behavior from data, the known physics of memory, plasticity, damage, and rate dependence can be embedded into the architecture while neural learning identifies the components that remain unknown. This combination of mechanistic structure and data-driven adaptability provides a promising foundation for the next generation of interpretable neural models for nonlinear structural and material systems.

## References

*Farrokh, M., Dizaji, M. S., & Joghataie, A. (2015). Modeling hysteretic deteriorating behavior using generalized Prandtl neural network. Journal of Engineering Mechanics, 141(8), 04015024.*

*Joghataie, A., & Shafiei Dizaji, M. (2016). Neuroplasticity in dynamic neural networks comprised of neurons attached to adaptive base plate. Neural Networks, 75, 77–83.*

*Farrokh, M., & Dizaji, M. S. (2016). Adaptive simulation of hysteresis using neuro-Madelung model. Journal of Intelligent Material Systems and Structures, 27(13), 1713–1724.*

*Joghataie, A., & Farrokh, M. (2008). Dynamic analysis of nonlinear frames by Prandtl neural networks. Journal of Engineering Mechanics, 134(11), 961–969.*

*Farrokh, M., & Joghataie, A. (2014). Adaptive modeling of highly nonlinear hysteresis using Preisach neural networks. Journal of Engineering Mechanics, 140(4), 06014002.*

*Joghataie, A., & Oliyan Torghabehi, O. (2014). Simulating dynamic plastic continuous neural networks by finite elements. IEEE Transactions on Neural Networks and Learning Systems, 25(8), 1583–1587.*

*Farrokh, M., Dizaji, F. S., & Dizaji, M. S. (2022). Hysteresis identification using extended Preisach neural network. Neural Processing Letters, 54, 1523–1547.*